# Particle Swarm Optimization: Fundamental Study and its Application to Optimization and to Jetty Scheduling Problems

J. Sienz[1] and M. S. Innocente[1]
[1]ADOPT Research Group,
School of Engineering
Swansea University,
Swansea, UK

## Abstract

The advantages of evolutionary algorithms with respect to traditional methods have been greatly discussed in the literature. While particle swarm optimizers share such advantages, they outperform evolutionary algorithms in that they require lower computational cost and easier implementation, involving no operator design and few coefficients to be tuned. However, even marginal variations in the settings of these coefficients greatly influence the dynamics of the swarm. Since this paper does not intend to study their tuning, general-purpose settings are taken from previous studies, and virtually the same algorithm is used to optimize a variety of notably different problems. Thus, following a review of the paradigm, the algorithm is tested on a set of benchmark functions and engineering problems taken from the literature. Later, complementary lines of code are incorporated to adapt the method to combinatorial optimization as it occurs in scheduling problems, and a real case is solved using the same optimizer with the same settings. The aim is to show the flexibility and robustness of the approach, which can handle a wide variety of problems.

## 1  Introduction

The characteristics of the objective variables, the function to be optimized and the constraint functions severely restrict the applicability of traditional optimization algorithms. The variables and both the objective and constraint functions must comply with a number of requirements for a given traditional method to be applicable. Furthermore, traditional methods are typically prone to converge towards local optima. By contrast, population-based methods such as evolutionary algorithms (EAs) and particle swarm optimization (PSO) are general-purpose optimizers, which are able to handle different types of variables and functions with few or no adaptations. Besides, although finding the global optimum is not guaranteed, they are able to escape poor local optima by evolving a population of interacting individuals which profit from information acquired through experience, and use stochastic weights or operators to introduce new responses. The lack of limitations to the features of the variables and functions that model the problem enable these methods to handle models whose high complexity does not allow traditional, deterministic, analytical approaches. While the advantages of PSO and EAs with respect to traditional methods are roughly the same, the main advantages of PSO when compared to EAs are its lower computational cost and easier implementation. Regarding their drawbacks, both these methods require higher computational effort, some constraint-handling technique incorporated, and find it hard to handle equality constraints.

Population-based methods like EAs and PSO are considered *modern heuristics* because they are not designed to optimize a given problem deterministically but to carry out some procedures that are not directly related to the optimization problem. Optimization occurs without evident links between the implemented technique and the resulting optimization process. They are also viewed as *Artificial Intelligence* (AI) *techniques* because their ability to optimize is an emergent property that is not specifically intended, and therefore not implemented in the code. Thus, the problem per se is not analytically solved, but artificial-intelligent entities are implemented, which are expected to find a solution themselves. In particular, *Swarm Intelligence* (SI) is the branch of AI concerned with the study of the collective behaviour that emerges from decentralized and self-organized systems. It is the property of a system whose individual parts interact locally with one another and with their environment, inducing the emergence of coherent global patterns that the individual parts are unaware of. PSO is viewed as one of the most prominent SI-based methods. Either modern heuristics or AI-based optimizers, these methods are not deterministically designed to optimize. EAs perform some kind of artificial evolution, where individuals in a population undergo simulated evolutionary processes which results in the maximization of a fitness function, resembling biological evolution. Likewise, PSO consists of a sort of simulation of a social milieu, where the ability of the population to optimize its performance emerges from the cooperation among individuals.

Although the basic particle swarm optimizer requires the tuning of a few coefficients only, even marginal variations in their values have a strong impact on the dynamics of the swarm. The general-purpose settings used throughout this paper were taken from previous studies (see Innocente [2]) because the objective is not to study their tuning but to demonstrate that virtually the same algorithm is able to cope with a variety of problems that would require different traditional methods to be solved. Thus, this paper intends to introduce the PSO method, pose the optimization and scheduling problems, and solve them with essentially the same algorithm. Additional code is required to turn the continuous search algorithm into a scheduler.

## 2 Mathematical optimization

For problems where the quality of a solution can be quantified in a numerical value, *optimization* is the process of seeking the permitted combination of variables that optimizes that value. Thus, different combinations of *variables* allow trying different candidate solutions, the *constraints* limit the valid combinations, and the *optimality criterion* allows differentiating better from worse. The suitability of traditional methods is limited by the nature of the variables, and by requirements that the objective and constraint functions must comply with.

The process of problem-solving consists of two major steps: developing a model of the problem, and finding a solution to it. The behaviour of an existing system can be sometimes analyzed by observation, but the number of situations analyzed is necessarily limited. When different alternatives need to be considered −for instance for optimization processes−, the development of a model becomes essential. Either mathematical or physical (*prototype*), any model is just an interpretation of a problem, which always implies simplifications. This work is concerned with mathematical models, so that the plain words in which real problems are posed must be turned into mathematical language (*formulation of the problem*). The question is whether to introduce numerous simplifications so that the available methods are able to solve the model, or to develop a model with higher fidelity and approximate the solving techniques. Traditional methods require specific characteristics of the equations involved, forcing the model to take the corresponding simplifying assumptions to suit the solver. This results in an exact solution of an approximate model. The alternative is to develop more precise models, and then attempt to solve them using a set of toolboxes that do not severely limit the model. This results in an approximate solution of a more precise model (e.g. finite element method, EAs and PSO). The second approach usually outperforms the first for complex, real-world problems.

While the *objective function* relates the real problem to the model, the *cost function* (also *evaluation function*) is the function to be minimized. The *problem variables* are called *object variables*, or just *variables*. Since *parameters* might stand for *variables*, the *parameters* of an algorithm are referred to as *coefficients* in this work.

## 2.1 Types of optimization problems

Optimization problems can be differentiated according to the models that represent them. Thus, they can be classified as *linear* or *non-linear*; *continuous*, *discrete* or *mixed-integer*; *convex* or *non-convex*; *differentiable* or *non-differentiable*, *smooth* or *non-smooth*, etc. It is interesting to classify them here into *continuous* and *discrete* because the PSO method was originally designed for continuous search-spaces, regardless of whether the problem is or is not *linear*, *convex*, *differentiable*, or *smooth*.

### 2.1.1 Continuous optimization

The variables in continuous problems can take real values. Since there are infinite real numbers, trying every possible solution is not an option, and the problem has no solution when the landscape that represents it is random. However, there is almost always a function that models the landscape –not necessarily continuous−, which returns information about the systematic relationships between its points.

### 2.1.2 Discrete optimization

A discrete problem has a finite number of possible solutions, although such number becomes intractable for real-world problems. Again, trying them all is out of question. Some branches of discrete optimization are binary optimization, linear programming, integer programming and combinatorial optimization. Due to the application of the PSO method to scheduling problems, the latter is of special interest here. The aim in combinatorial optimization is to find the optimal arrangement of the elements that optimizes a result. The simplest object variable is a list, where the different solutions consist of its permutations. A classical example is the travelling salesman problem. Traditional methods for combinatorial problems are sequential.

## 2.2 General optimization problem

Let $S$ be the search-space, and $\mathcal{F} \subseteq S$ its feasible part. A minimization problem consists of finding $\hat{\mathbf{x}} \in \mathcal{F}$ such that $f(\hat{\mathbf{x}}) \leq f(\mathbf{x}) \ \forall \mathbf{x} \in \mathcal{F}$, where $f(\hat{\mathbf{x}})$ is a global minimum and $\hat{\mathbf{x}}$ its location. The problem can be formulated as in Equation (1):

$$\text{Minimize } f(\mathbf{x}), \quad \text{subject to } \begin{cases} g_j(\mathbf{x}) \geq 0 & ; \quad j = 1, \ldots, q \\ g_j(\mathbf{x}) = 0 & ; \quad j = q+1, \ldots, m \end{cases} \quad (1)$$

where $\mathbf{x} \in S$ is the vector of object variables; $f(\cdot): S \to \mathcal{E}$ is the cost function; $g_j(\cdot)$ is the $j^{\text{th}}$ constraint function; and $S \subseteq \mathcal{R}^n \land \mathcal{E} \subseteq \mathcal{R}$ for continuous problems ($\mathcal{R}$ is the set of real numbers).

### 2.2.1 The objective and the cost functions

The *objective* of an optimization problem is given in plain words. Multi-objective problems seek to achieve several −typically conflicting− objectives. The *formulation* of the objective is called the *objective function*, whose output is a measure of the fulfilment of the objective(s). The *cost function* is the function to be minimized, and maps the output of the *objective function* to the real numbers so that the minimum matches the best fulfilment. The *cost* and the *objective* functions may coincide; be proportional; or there can be an arbitrarily complex mapping between them.

### 2.2.2 Constraints

Although the constraints allow concentrating the search into limited areas, each potential solution must be verified to comply with them. Hence the problem usually becomes harder to solve than its unconstrained counterpart. The most appropriate technique to handle a constraint depends on the type of constraint. Some common types of constraints are as follows:

- *Inequality constraint*: function of the object variables that must be greater than or equal to a constant, as shown in Equation (1).
- *Equality constraint*: function of the object variables that must be equal to a constant, as shown in Equation (1).
- *Boundary constraint*: instance of inequality constraints, consisting of functions that define boundaries that contain the feasible space (if the boundary constraints are given by a hyper-rectangle, they are also called *interval* or *side constraints*).

While a constrained optimization problem was defined as the problem of finding the combination of variables that minimizes the cost function and satisfies all constraints, real-world problems sometimes do not lend themselves to such strict conditions. Frequently, all the constraints cannot be strictly satisfied simultaneously, and the problem turns into finding a trade-off between minimizing the cost function and minimizing the constraints' violations. Thus, a constraint is said to be *hard* if it does not admit any degree of violation, and *soft* is there is some given tolerance.

**Constraint-handling techniques**

Constraint-handling techniques are usually classified into groups, although such classifications are not unified. Besides, while some methods are sometimes regarded as belonging to one class, the boundaries are frequently blurry, and the memberships not unique. The focus here is on those techniques suitable for PSO.

Rejection of infeasible solutions

This is a *penalization* method, sometimes referred to as the *death penalty* due to the elimination of infeasible solutions. The algorithm does not need to evaluate infeasible solutions, thus saving computational effort. The method is suitable when $\mathcal{F}$ is convex, and when it constitutes an important part of $\mathcal{S}$. It presents some problems like the fact that the solution(s) must be initialized within $\mathcal{F}$, and cannot be placed within infeasible regions on the way towards other disjointed regions of $\mathcal{F}$.

Penalization of infeasible solutions

The evaluation of infeasible solutions might be useful to guide the search towards more promising areas. Thus, the value of the *cost function* is increased for infeasible solutions as in Equation (2), and the problem is treated as if it was unconstrained.

$$fp(\mathbf{x}) = f(\mathbf{x}) + Q(\mathbf{x}) \qquad (2)$$

where $f(\mathbf{x})$ and $fp(\mathbf{x})$ are the cost and penalized cost functions, respectively, while $Q(\mathbf{x})$ is the overall penalization due to the infeasibility of solution $\mathbf{x}$.

Penalties are usually not fixed but linked to the amount of infeasibility. A classical penalization for the $j^{\text{th}}$ constraint violation is shown in Equation (3):

$$f_j(\mathbf{x}) = \begin{cases} \max\{0, g_j(\mathbf{x})\} & ; \quad 1 \leq j \leq q \\ g_j(\mathbf{x}) & ; \quad q < j \leq m \end{cases} \quad (3)$$

Thus, the function to be optimized could be as follows (additive penalization):

$$fp(\mathbf{x}) = f(\mathbf{x}) + \sum_{j=1}^{m} \lambda_j^{(t)} \cdot (f_j(\mathbf{x}))^\alpha \quad (4)$$

where the penalization coefficients $\lambda_j^{(t)}$ and $\alpha$ may be constant, dynamic, or adaptive. Typically, $\lambda_j^{(t)}$ is set to high and $\alpha$ to small values. A high penalization might lead to infeasible regions of $S$ not being explored, thus converging to a non-optimal but feasible solution. A low penalization might lead to the system evolving solutions that are violating constraints but present themselves as having lower costs than feasible solutions. The definition of the penalty functions is not trivial, and it plays a critical role in the performance of the algorithm. A wide variety of penalization methods can be found in the literature, including the multiplicative ones.

Preserving feasibility

Infeasible solutions are ignored, and candidate solutions quickly pulled back towards $F$. While it *rejects infeasible solutions*, it is considered here a *preserving feasibility* method in agreement with Hu et al. [3]. Its drawbacks are that the solution(s) must be initialized within $F$; it is inefficient in handling small and disjointed feasible spaces; and cannot handle equality constraints. Its advantage is that it only needs a slight adaptation from the unconstrained algorithm, without coefficients to be tuned.

Cut off at the boundary

When a solution is moved to an infeasible location, its displacement vector is simply cut off. Either the solution is relocated on the boundary the nearest possible to the attempted new location, or the direction of the displacement vector is kept unmodified. Typically, the first alternative is preferred.

Bisection method

The *cut off at the boundary* technique works well when the boundaries are limited by interval constraints, and when the solution is located on the boundary. However, solutions might get trapped in the boundaries when the optimum is near them. Foryś et al. [4] proposed a *reflection technique* instead, while Innocente [2] proposed the *bisection method*: if the new position is infeasible, the displacement vector is split in halves, and the constraints verified. If the new position is still infeasible, the vector is split again, and so on. A feasible position will be found in time, unless the particle is already on the boundary. Hence an upper limit of iterations is set, and the particle keeps its position if no new feasible location can be found. Its drawbacks are that it cannot pass through infeasible space, and that the solution(s) must be initialized within $\mathcal{F}$. These drawbacks are shared with the *cut off at the boundary*, and to some extent with the *preserving feasibility* methods. Its main advantages are that it needs no adaptations to handle different inequality constraints, and its fast convergence.

Repair infeasible solutions

Engelbrecht [12] defines them as methods that "…allow particles to move into infeasible space, but special operators (methods) are then applied to either change the particle into a feasible one or to direct the particle to move towards feasible space. These methods usually start with initial feasible particles". Note that *particle* stands for *candidate solution*.

According to Engelbrecht [12], the *preserving feasibility* methods are those which ensure that adjustments to the particles do not violate any constraint. Hence the *cut off at the boundary* technique and the *bisection method* could be considered as *preserving feasibility* methods, while the *preserving feasibility* method discussed above could be considered as a method that *rejects infeasible solutions*.

## 2.3 Optimization methods

The important differences among the different types of optimization problems make it necessary to handle each type by means of completely different approaches. Optimization methods can be classified according to the **type of problems they are able to handle** (i.e. *linear*, *integer*, *mixed-integer*, *binary*, *quadratic*, or *non-linear*; *differentiable* or *non-differentiable*; *smooth* or *non-smooth*; *continuous* or *discrete*), or according to **the features of the algorithms** (i.e. *gradient-based* or *gradient-free*; *exact* or *approximate*; *deterministic* or *probabilistic*; *analytical* or *heuristic*, *single-based* or *population-based*). There are numerous different and often conflicting classifications. In addition, words like *heuristics* do not have a precise definition in this context. In general, it refers to techniques that do not guarantee to find anything, and are usually based on common sense, natural metaphors, or even methods whose behaviour is not fully understood. The classification offered in Figure 1 adheres to this definition. Since the membership of some paradigms to one or another class is sometimes unclear and conflicting, the criterion used is that if a *heuristic* qualifies as a *traditional method*, the latter prevails (e.g. *Tabu Search*). Purely deterministic methods are always viewed as traditional here, while probabilistic methods as heuristics.

*Modern heuristics* are general-purpose methods that will do well on most problems, although a problem-specific algorithm would be probably more efficient. They use techniques that are not directly related to optimization processes, heavily relying on stochastic operators. Their status of general-purpose algorithms together with their ability to escape local optima, ridges, and plateaus are their major advantages.

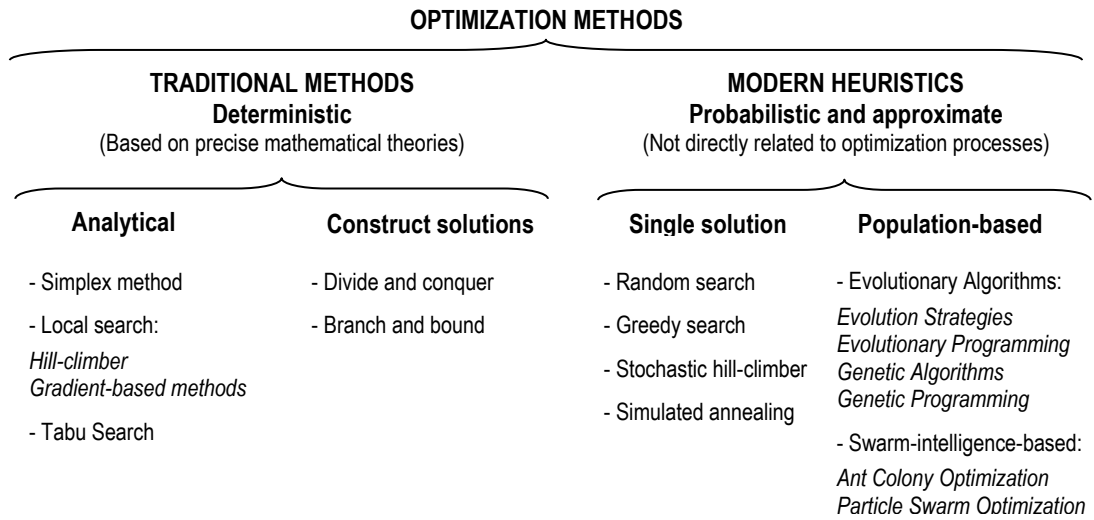

Figure 1: Classification of different types of optimization methods.

## 2.4 Population-based methods

Population-based methods have proven to be applicable to a variety of optimization problems, data mining, pattern recognition, classification, machine learning, scheduling, supply-chain management, etc. These outstandingly robust methods outperform traditional methods when dealing with problems of high complexity. They successively update a population of candidate solutions, thus performing a parallel exploration instead of the sequential exploration carried out by traditional methods, whose point-to-point search is usually unable to overcome local pathologies. In addition, these methods only require the cost function information to guide the search, without the need of auxiliary information such as gradients and Hessian matrices.

# 3 Particle Swarm Optimization

The original PSO method was designed by social-psychologist James Kennedy and electrical-engineer Russell C. Eberhart in 1995 [5]. The main algorithm was inspired by earlier bird flock simulations framed within the field of social psychology. Therefore, the method is closely related to other simulations of social processes and experimental studies in social psychology, while also having strong roots in both optimization and AI. To be more precise, it can be placed within the subfields *Computational Intelligence* (CI)[1] and/or *Artificial Life* (AL)[2]. Some of the experimental stud-ies in social psychology that influenced the method are Lewin's field theory; Gestalt theory; Sherif's and Asch's experiments; Latané's social impact theory; Bandura's no-trial learning; simulations of the spread of culture in a population, and of the behaviour of social animals such as bird flocks and social insects (refer to Kennedy et al. [1] or Innocente [2] for further reading). From an optimization point of view, it is a search method suitable for optimization problems whose solutions can be represented as points in an *n*-dimensional space. While the variables needed to be real-valued in the original version, binary and other discrete versions have also been developed (e.g. Kennedy et al. [1], Kennedy et al. [6], Clerc [7], Mohan et al. [8]).

---

[1] AI divides into different schools of thought with vague boundaries. Some suggest two mainstreams: **classical AI** refers to the *symbolic paradigm* and its derivatives, and **CI** to all other paradigms such as *artificial neural networks* (ANNs), *fuzzy systems*, EAs and PSO. Others include both the *symbolic* and the *connectionist* (i.e. ANNs) *paradigms* into **classical AI**, while **modern AI** comprises the rest.
[2] *Artificial life* (AL) encompasses paradigms based upon metaphors of the behaviour of biological organisms, such as *cellular automaton*, EAs and PSO. The aim is not to model biological life but to create artificial beings that exhibit degrees of aliveness and intelligence by exploiting principles underlying biological organisms, where evolution and the ability to learn are critical for their survival.

The PSO method relies on random coefficients to introduce creativity. It is a bottom-up approach in the sense that the system's intelligent behaviour emerges in a higher level than the individuals', evolving intelligent solutions without using the programmers' expertise on the subject matter. While this feature makes it difficult to understand the way optimization is actually performed, it shows astonishing robustness in handling many kinds of complex problems that it was not specifically designed for. However, it presents the disadvantage that its theoretical bases are extremely difficult to be understood deterministically. Nevertheless, considerable theoretical work has been carried out on simplified versions of the algorithm, extrapolated to the full version, and finally supported by experimental results (e.g. Trelea [9], Clerc et al. [10], Ozcan et al. [11]). Refer to Kennedy et al. [1], Engelbrecht [12] and Clerc [13] for a general, comprehensive review of the paradigm.

## 3.1 Continuous, unconstrained PSO

The PSO method was originated on the simulation of a simplified social milieu, where individuals (i.e. *particles*) were thought of as collision-proof birds. The population is usually referred to as the *swarm*, while the function to be minimized is called here *conflict* function due to the social-psychology metaphor that inspired the method. That is, each individual searches the space of beliefs, seeking the minimization of the *conflicts* among the beliefs it holds by using the information gathered by both its own experience and those of others. Individuals indirectly seek agreement by clustering in the space of beliefs, which is −broadly speaking− the result of all the individuals imitating their most successful peer(s), thus becoming more similar to one another as the search progresses. However, the clustering is delayed by their own previous successful experiences, which each individual is reluctant to disregard.

To summarize, while the emergent properties of the PSO algorithm result from local interactions among the particles in the swarm, the behaviour of a single individual particle can be summarized in three sequential processes:

- **Evaluation**. The particle evaluates its position in the environment, which is given by the associated value of the conflict function. Following the social psychology metaphor, this would stand for the conflict among its current set of beliefs.
- **Comparison**. Once the particle's position in the environment is evaluated, it is not straightforward to tell how good it is. Experiments and theories in social psy-chology suggest that humans judge themselves by comparing to others, thus telling better from worse rather than good from bad. Therefore, the particle compares the conflict among its current set of beliefs to those of its neighbours.
- **Imitation**. The particle imitates those whose performances are superior in some sense. In the basic PSO (B-PSO), only the most successful neighbour is imitated.

These three processes are implemented within PSO, where the only sign of individual intelligence is a small memory. The basic update equations are as follows:

$$v_{ij}^{(t)} = w \cdot v_{ij}^{(t-1)} + iw \cdot U_{(0,1)} \cdot \left(pbest_{ij}^{(t-1)} - x_{ij}^{(t-1)}\right) + sw \cdot U_{(0,1)} \cdot \left(gbest_{j}^{(t-1)} - x_{ij}^{(t-1)}\right) \quad (5)$$

$$x_{ij}^{(t)} = x_{ij}^{(t-1)} + v_{ij}^{(t)} \quad (6)$$

where $x_{ij}^{(t)}$ and $v_{ij}^{(t)}$ are the $j^{th}$ coordinate of the position and velocity, respectively, of particle $i$ at time-step $t$; $U_{(0,1)}$ is a random number from a uniform distribution in the range [0,1] (resampled anew every time it is referenced); $w$, $iw$ and $sw$ are the inertia, individuality, and sociality weights, respectively; $pbest_{ij}^{(t-1)}$ and $gbest_{j}^{(t-1)}$ are the $j^{th}$ coordinate of the best position found by particle $i$ and by any particle in the swarm, respectively, by time-step $(t-1)$.

As shown in Equations (5) and (6), there are three coefficients in the basic algorithm which rule the dynamics of the swarm: the *inertia* ($w$), the *individuality* ($iw$), and the *sociality* ($sw$) *weights*. The $iw$ and the $sw$ are sometimes referred to as the *learning weights*, while their aggregation is called here the *acceleration weight* ($aw$).

Thus, the performance of a particle in its current position is **evaluated** in terms of the *conflict* function. In order to decide upon its next position, the particle **compares** its current conflict to those associated to both, its own and its neighbours' best previous experiences (*pbest* and *gbest*, respectively). Finally, the particle **imitates** the best experience of its most successful neighbour, but without disregarding its own.

The relative importance given to $iw$ and $sw$ results in the particles exhibiting more self-confident or more conformist behaviour, while the random weights introduce creativity into the system: since they are resampled anew for each time-step, for each particle, for each component, and for each term in Equation (5), the particles display uneven trajectories that allow better exploration of the search-space. In addition, resampling them anew for the individuality and the sociality terms −together with setting $iw = sw$− leads to the particles alternating self-confident and conformist behaviour. Typically, $iw = sw = 2$ (i.e. $aw = 4$). Every particle also tends to keep its current velocity, where the strength of this tendency is governed by $w$. The relative importance between $w$ and $aw$ results in more explorative or more exploitative behaviour of the swarm.

In the original PSO, Kennedy et al. [5] did not consider $w$ (i.e. $w = 1$), and suggested setting $iw = sw = 2$. However, the particles tended to diverge rather than clus-ter, and the swarm appeared to perform a so-called *explosion*. It was found that if the components of the particles' velocities were bounded, the explosion could be controlled and the particles ended up clustering around a solution. Later, aiming to control the explosion, Shi et al. [14] proposed the incorporation of the inertia weight ($w$).

Based on the B-PSO, countless variations can be thought of. For instance, Clerc et al. [13] proposed the use of a constriction factor −which can be viewed as an instance of the $w$− to control the explosion of the original PSO and guarantee convergence; different settings of the coefficients can be thought of (which can be constant, deterministically dynamic, or adaptive); the coefficients and velocity constraint can be scalars or impose different behaviour for different dimensions; the neighbourhoods can be *global* or *local*; different neighbourhoods' architecture can be designed; the best experiences can be updated after all the particles' positions in the neighbourhood have been updated or every time a particle's position is updated, etc. As to the discrete version (D-PSO), alternatives like rounding off the particles' positions to match the discrete values and the binary PSO (b-PSO) can be thought of.

**About PSO as AI-based optimization method**

AI is a polemical field due to controversies in the very definition of intelligence. Countless definitions can be found, most of them involving concepts like exhibiting verbal, mathematical and problem-solving abilities; capacity to think and reason; and/or ability to acquire, store and apply knowledge. While these definitions enumerate qualities, it can be argued that those are symptoms of intelligence. If intelligence is to be defined by enumerating its symptoms, all of them should be specified. If intelligence could be unquestionably defined, it would be reasonable to assert that AI is the intelligence exhibited by any artificial entity. However, definitions for biological intelligence and AI frequently differ. Fogel et al. [15] claim that a proper definition should apply to humans and machines equally well, and suggested that "intelligence may be defined as the capability of a system to adapt its behaviour to meet its goals in a range of environments" (quoted in [15]). Thus, it makes sense to speak of biological and artificial entities exhibiting degrees of intelligence, as different features of intelligence −and to different extents− are displayed.

The swarm in the PSO method displays problem-solving ability; each particle learns and profits from its own and others' previous experiences (stored in a small memory); and most importantly, the swarm *adapts its behaviour to meet its goals in a range of environments*, which gives the algorithm the status of a *general-purpose* problem-solver. Therefore, it can be claimed to exhibit some degree of intelligence.

### 3.2 Binary, unconstrained PSO

The b-PSO was originally proposed by Kennedy at al. [6] as a variation of the continuous version. The search-space consists of an $n$-dimensional binary hyper-cube; the particles are represented by bit-strings; and the conflict function $c : \{0,1\}^n \to \mathcal{R}$. The basic update equations are now as follows:

$$v_{ij}^{(t)} = v_{ij}^{(t-1)} + iw \cdot U_{(0,1)} \cdot \left(pbest_{ij}^{(t-1)} - x_{ij}^{(t-1)}\right) + sw \cdot U_{(0,1)} \cdot \left(gbest_{j}^{(t-1)} - x_{ij}^{(t-1)}\right) \quad (7)$$

$$p_{ij}^{(t)} = \frac{1}{1 + e^{-v_{ij}^{(t)}}} \in [0,1] \subset \mathcal{R} \quad (8)$$

$$x_{ij}^{(t)} = \begin{cases} 1 & \text{if} \quad U_{(0,1)} < p_{ij}^{(t)} \\ 0 & \text{otherwise} \end{cases} \quad (9)$$

where $p_{ij}^{(t)}$ stands for the probability of a bit adopting state 1. That is, the probability that individual *i* has of holding belief *j* at time-step *t*. The other parameters in Equation (7) are the same as in Equation (1). The velocities can still take real values.

### 3.3 Constrained PSO

The *additive penalization* as posed in Equation (4), the *preserving feasibility* and the *bisection* methods were implemented. The penalization coefficients were not tuned but intuitively set as follows: $\lambda = 10^6$ and $\alpha = 2$. The study of their tuning and self-adaptation is beyond the scope of this paper, as well as the adaptations of the *preserving feasibility* and *bisection* methods to handle equality constraints.

After extensive study of the impact of the three basic coefficients in Equations (5) on the behaviour of the swarm (see Innocente [2]), it was concluded that there is no general-purpose setting that provides together the abilities to fine-cluster and to escape poor sub-optimal solutions. The strategy adopted was to sub-divide the swarm in three sub-swarms whose set of coefficients grant them complementary abilities.

## 4 Benchmark optimization problems

It is important to keep in mind that the objective here is not fast convergence but to show that the same algorithm with the same settings can handle different problems.

### 4.1 First suite: unconstrained benchmark functions

The proposed general-purpose settings have to be tested on a suite of unconstrained benchmark problems, so that their evaluation is independent from the constraint-handling method. The suite is that used by Eberhart et al. [16] and Trelea [9] for comparison. Refer to their work for the equations and stopping criteria.

Typically, termination conditions in the literature consist of a maximum number of iterations and a given function target, only applicable for problems whose solution is known. The ones implemented in our solver are based on those studied by Innocente [2], suitable for real-world problems. However, in order to compare our results to those obtained by Eberhart et al. [16] and Trelea [9], the attainment of function targets terminates the search, the maximum number of time-steps is set to 10000, and the swarm size equals 30. A global (GP-GPSO) and a local version with 3-particle neighbourhood (GP-LPSO) are tested, and the results shown in Table 1.

| Function | Optimizer | Time-steps to meet function goal | | Success rate | Expected function evaluations to meet function goal | Average solution found | Best solution found | Time-steps | Function Evaluations |
|---|---|---|---|---|---|---|---|---|---|
| | | Average | Minimum | | | | | | |
| Sphere | Trelea Set 1 [9] | 344.00 | 266 | 1.00 | 1.03E+04 | - | - | - | - |
| | Trelea Set 2 [9] | 395.00 | 330 | 1.00 | 1.19E+04 | - | - | - | - |
| | Eberhart et al. [16] | 529.65 | 495 | 1.00 | 1.59E+04 | - | - | - | - |
| | GP-GPSO | 462.40 | 383 | 1.00 | 1.39E+04 | 1.42E-120 | 9.14E-134 | 1.00E+04 | 3.00E+05 |
| | GP-LPSO | 5977.75 | 4177 | 1.00 | 1.79E+05 | 2.36E-04 | 1.54E-07 | 1.00E+04 | 3.00E+05 |
| Rosenbrock | Trelea Set 1 [9] | 614.00 | 239 | 1.00 | 1.84E+04 | - | - | - | - |
| | Trelea Set 2 [9] | 900.00 | 298 | 1.00 | 2.70E+04 | - | - | - | - |
| | Eberhart et al. [16] | 668.75 | 402 | 1.00 | 2.01E+04 | - | - | - | - |
| | GP-GPSO | 679.35 | 336 | 1.00 | 2.04E+04 | 7.58E+00 | 4.34E-04 | 1.00E+04 | 3.00E+05 |
| | GP-LPSO | 6262.00 | 4840 | 0.30 | 6.26E+05 | 1.69E+02 | 4.02E+01 | 1.00E+04 | 3.00E+05 |
| Rastrigin | Trelea Set 1 [9] | 140.00 | 104 | 0.90 | 4.67E+03 | - | - | - | - |
| | Trelea Set 2 [9] | 182.00 | 123 | 0.95 | 5.75E+03 | - | - | - | - |
| | This paper Set 1 | 102.53 | 67 | 0.95 | 3.24E+03 | 6.78E+01 | 3.48E+01 | 1.00E+04 | 3.00E+05 |
| | This paper Set 2 | 156.40 | 106 | 1.00 | 4.69E+03 | 6.17E+01 | 3.58E+01 | 1.00E+04 | 3.00E+05 |
| | Eberhart et al. [16] | 213.45 | 161 | 1.00 | 6.40E+03 | - | - | - | - |
| | GP-GPSO | 211.50 | 117 | 1.00 | 6.35E+03 | 2.61E+01 | 9.95E+00 | 1.00E+04 | 3.00E+05 |
| | GP-LPSO | 2802.05 | 202 | 0.90 | 9.34E+04 | 6.92E+01 | 5.02E+01 | 1.00E+04 | 3.00E+05 |
| Griewank | Trelea Set 1 [9] | 313.00 | 257 | 0.90 | 1.04E+04 | - | - | - | - |
| | Trelea Set 2 [9] | 365.00 | 319 | 0.90 | 1.22E+04 | - | - | - | - |
| | Eberhart et al. [16] | 312.60 | 282 | 1.00 | 9.38E+03 | - | - | - | - |
| | GP-GPSO | 601.53 | 362 | 0.95 | 1.90E+04 | 4.00E-02 | 0.00E+00 | 1.00E+04 | 3.00E+05 |
| | GP-LPSO | 5860.00 | 3724 | 0.80 | 2.20E+05 | 4.67E-02 | 1.60E-04 | 1.00E+04 | 3.00E+05 |
| Schaffer f6 | Trelea Set 1 [9] | 161.00 | 74 | 0.75 | 6.44E+03 | - | - | - | - |
| | Trelea Set 2 [9] | 350.00 | 102 | 0.60 | 1.75E+04 | - | - | - | - |
| | Eberhart et al. [16] | 532.40 | 94 | 1.00 | 1.60E+04 | - | - | - | - |
| | GP-GPSO | 694.44 | 142 | 0.90 | 2.31E+04 | 3.70E-04 | 0.00E+00 | 1.00E+04 | 3.00E+05 |
| | GP-LPSO | 779.95 | 101 | 1.00 | 2.34E+04 | 0.00E+00 | 0.00E+00 | 1.00E+04 | 3.00E+05 |

Table 1: Summary of the results obtained for the suite of unconstrained benchmark functions. The settings in Trelea Set 2 [9] and Eberhart et al. [16] are the same, while those in Trelea [9] were replicated in "This paper" for the Rastrigrin function.

Eberhart et al. [16] and Trelea [9] implemented settings that favour fine-clustering and convergence, so that their optimizers quickly achieve the conveniently set function targets. However, premature clustering takes place for time-extended searches −especially for multi-modal problems such as the Rastrigrin function−, as studied by Innocente [2]. In contrast, further improvement is possible for our optimizers because a third of the swarm has the ability to escape local optima. In order to illustrate this, the function targets were removed and our optimizers also tested for the whole 10000 time-steps. The settings in Eberhart et al. [16] and Trelea [9] were replicated and also tested along 10000 time-steps (on the Rastrigrin function only), showing their premature clustering. The results are shown in Table 1, where the white background indicates those corresponding to the function targets (see Trelea [9]), and the shaded background the results corresponding to the whole 10000 time-steps. The coefficients in *Trelea Set 2 [9]*, *Eberhart et al. [16]* and *This paper Set 2* coincide. The same is true for *Trelea Set 1 [9]* and *This paper Set 1*.

Although its coefficients were not aimed at fast convergence, the number of time-steps that the GP-PSO require to meet the targets for the first three functions is comparable to those of Eberhart et al. [16] and Trelea Set 2 [9], while its success rate outperforms those in Trelea [9] overall. As to the neighbourhood topology, the objective of the local versions is to help avoid premature clustering. Therefore, it is not surprising that the GP-LPSO –whose both coefficients and neighbourhood topology delay convergence– require many more time-steps to meet the targets. It is important to remark that the failures to achieve such goals along the permitted 10000 time-steps is due to the delay in the particles' clustering rather than to premature convergence as in the cases of Eberhart et al. [16] and Trelea [9]. This can be inferred from the fact that the improvement of solutions does not stagnate, and by comparing the evolution of the measures of clustering between the GP-GPSO and the GP-LPSO (not included here). Thus, the local version delays the loss of diversity and hence convergence, keeping the ability to escape local optima for longer. This is especially convenient for multi-modal problems such as the Rastrigrin and Schaffer f6 functions. The former is 30-dimensional in the test suite, which results in the permitted 10000 time-steps being insufficient for the particles to fine-cluster, whereas the latter is only 2-dimensional. Thus, the poor solutions found by the GP-LPSO for the first four 30-dimensional functions notably improve for searches extended beyond the permitted 10000 time-steps, while it finds the exact solutions for the 2-dimensional Schaffer f6 function in every run (see last row in Table 1).

The settings in *Trelea Set 1 [9]* were replicated in *This paper Set 1*; those corresponding to both *Trelea Set 1 [9]* and *Eberhart et al. [16]* in *This paper Set 2*; and both tested on the Rastrigrin function for the whole 10000 time-steps. It can be seen that they meet the target in less than 160 time-steps on average, their best solution out of 20 runs is around 35, and their average solution over 60. In contrast, the GP-GPSO meets the target in over 210 time-steps on average, but its best solution after 10000 time-steps is notably better (9.95), while its average solution in 20 runs (26) notably smaller even than the best found by the other optimizers! The GP-LPSO is far from converging, and improvement still occurs even after 30000 time-steps.

In summary, settings that favour convergence and a global version are convenient for expensive function evaluations (*fes*) and for convex functions, while settings that do not favour convergence and local versions are convenient when exploration needs to be greatly improved (e.g. multi-modal functions, heavily constrained problems and disjointed feasible spaces). The GP-GPSO is a robust, general-purpose trade-off.

### 4.2  Second suite: constrained benchmark functions

The optimizer also needs to be tested on a suite of constrained functions because such constraints affect the normal behaviour of the system.

### 4.2.1 Pressure vessel problem

The problem −as posed by Hu et al. [17]− is mixed-discrete (m-d), where $x_1$ and $x_2$ are integer multiples of 0.0625, while $x_3$ and $x_4$ are continuous. Thus, a swarm composed of 20 particles, 10000 time-steps and 11 runs per test are implemented, as in Hu et al. [17]. The results are also compared to those obtained by Coello Coello [18] in Table 2 and Table 3. Since Hu et al. [17] and Coello Coello [18] reported their results as the best among other authors', such comparisons are omitted here.

| PRESSURE VESSEL | | 1 neighbour | | 2 neighbours | | Global | |
|---|---|---|---|---|---|---|---|
| | | Real | Penalized | Real | Penalized | Real | Penalized |
| Penalization | Best | 6033.618818 | 6050.947591 | 6040.076927 | 6049.857978 | **6040.078181** | **6049.857964** |
| | Mean | 6173.760461 | 6179.567556 | 6127.535794 | 6135.206712 | **6398.899925** | **6411.395265** |
| Preserving feasibility | Best | **6066.259215** | - | 6090.526202 | - | 6090.526202 | - |
| | Mean | **6175.946584** | - | 6328.738977 | - | 6547.361378 | - |
| Bisection | Best | 6090.526202 | - | **6059.714335** | - | 6090.526202 | - |
| | Mean | 6503.193272 | - | **6476.062984** | - | 6539.554416 | - |
| Hu et al. [17] | Best | - | - | **6059.131296** | - | - | |
| Coello Coello [18] | Best | 6288.744500 | | | | | |
| | Mean | 6293.843232 | | | | | |

Table 2: Minima found by our optimizer with three constraint-handling methods and three neighbourhood sizes, and by two other authors for the m-d pressure vessel problem.

| Pressure Vessel | $x_1$ | $x_2$ | $x_3$ | $x_4$ |
|---|---|---|---|---|
| Penalization | 0.812500 | 0.437500 | 42.260491 | 174.638894 |
| Preserving feasibility | 0.812500 | 0.437500 | 42.044793 | 177.302575 |
| Bisection | 0.812500 | 0.437500 | 42.098446 | 176.636596 |
| Hu et al. [17] | 0.812500 | 0.437500 | 42.098450 | 176.636600 |
| Coello Coello [18] | 0.812500 | 0.437500 | 40.323900 | 200.000000 |

Table 3: Coordinates of the minima found by our optimizer and by two other authors for the m-d pressure vessel problem.

He et al. [19] proposed a hybrid PSO (simulating annealing incorporated), and tested it on this problem using 250 particles, 81000 *fes*, and 30 runs for the statistics. They reported to have outperformed several other authors. Hence our optimizer is tested again using the same number of particles and *fes*, and the results compared only to theirs (see Table 4). It must be noted that the general-purpose coefficients used here were designed for a swarm of 20-50 particles, 1000-30000 time-steps, and 1 to 30-dimensional problems. While the appropriate swarm-size, degree of locality, and number of object variables are clearly related, the study of such relationships is beyond the scope of this paper. It is fair to mention, however, that fast convergence results not only from appropriate coefficients' settings but also from small swarm sizes, highly connected neighbourhoods, and constraint-handling methods that quickly decrease the particles' momentum. Notice that the swarm size and *fes* used by He et al. [19] result in a more parallel search and shorter evolution. Therefore, coefficients' settings that favour convergence would be probably more appropriate.

| PRESSURE VESSEL | | 2 neighbours | | Global | |
|---|---|---|---|---|---|
| | | Real | Penalized | Real | Penalized |
| Penalization | Best | **6036.019786** | **6050.289209** | 6073.148086 | 6081.860801 |
| | Mean | **6054.780217** | **6064.294442** | 6309.524952 | 6317.137956 |
| Preserving feasibility | Best | 6061.308346 | - | **6059.714335** | - |
| | Mean | 6092.492880 | - | **6422.404101** | - |
| Bisection | Best | **6059.714335** | - | 6059.714335 | - |
| | Mean | **6073.456533** | - | 6360.990084 | - |
| He et al. [19] | Best | - | - | **6059.714300** | - |
| | Mean | - | - | **6099.932300** | - |

Table 4: Minima found by our optimizer with three constraint-handling methods and two neighbourhood sizes, and by He et al. [19] for the m-d pressure vessel problem.

Although de Freitas Vaz et al. [20] claimed that their optimizer returned better results than those reported by Hu et al. [17], the comparison is invalid because they posed the problem as continuous. The same is true in Foryś et al. [4] and Innocente [2]. In order to compare our results to those of de Freitas Vaz et al. [20], 30 analyses are performed using their settings: continuous variables and 879000 *fes*. The results are also compared to those of Foryś et al. [4] for reference (see Table 5 and Table 6).

| PRESSURE VESSEL | | 2 neighbours | | Global | |
|---|---|---|---|---|---|
| | | Real | Penalized | Real | Penalized |
| Penalization | Best | 5851.887101 | 5879.828359 | **5854.933947** | **5870.123975** |
| | Mean | 5872.459619 | 5886.238461 | **5854.933953** | **5870.123975** |
| Preserving feasibility | Best | 5888.171310 | - | **5886.483117** | - |
| | Mean | **5944.336034** | - | 6130.527590 | - |
| Bisection | Best | **5887.586796** | - | 5918.329579 | - |
| | Mean | **6046.356830** | - | 6362.007338 | - |
| de Freitas Vaz et al. [20] | Best | - | - | **5885.330000** | - |
| Forys et al. [4] | Best | | | 5885.490000 | |

Table 5: Minima found by our optimizer with three constraint-handling methods and two neighbourhood sizes, and by two other authors for the pressure vessel problem.

| Pressure Vessel | $x_1$ | $x_2$ | $x_3$ | $x_4$ |
|---|---|---|---|---|
| Penalization | 0.774549 | 0.383204 | 40.319619 | 200.000001 |
| Preserving feasibility | 0.778841 | 0.384982 | 40.354458 | 199.515579 |
| Bisection | 0.779484 | 0.385300 | 40.387763 | 199.053548 |
| de Freitas Vaz et al. [20] | 0.778169 | 0.384649 | 40.319600 | 200.000000 |
| Forys [4] | 0.778300 | 0.384700 | 40.324400 | 199.933000 |

Table 6: Coordinates of the minima found by our optimizer and by two other authors for the pressure vessel problem.

Given that a swarm of 30 particles is implemented here, the search is carried out throughout 29300 time-steps, so as to match the number of *fes* used in de Freitas Vaz et al. [20]. This long search combined with this small swarm and a low-dimensional problem gives time for the local versions to achieve a reasonable degree of clustering, although improvement still does not stagnate for local versions with penalization and preserving feasibility methods by the time the search is ended.

### 4.2.2 Welded beam problem

For the formulation of this problem, refer to Hu et al. [17] or Coello Coello [18]. In order to compare our results to those obtained by Hu et al. [17], their settings are replicated, and also compared to those of other authors for reference despite the algorithmic details being different. Some of the results are shown in Table 7.

| WELDED BEAM | | 1 neighbour | | 2 neighbours | | Global | |
|---|---|---|---|---|---|---|---|
| | | Real | Penalized | Real | Penalized | Real | Penalized |
| Penalization | Best | 1.72485136 | 1.72485185 | **1.72485135** | **1.72485185** | 1.72485136 | 1.72485185 |
| | Mean | 1.74076752 | 1.74076893 | **1.72485139** | **1.72485185** | 1.73616402 | 1.73616456 |
| Preserving feasibility | Best | 1.72485231 | 1.72485231 | **1.72485231** | - | 1.72485231 | 1.72485231 |
| | Mean | 1.72487563 | 1.72487563 | **1.72485234** | - | 1.73528547 | 1.73528547 |
| Bisection | Best | 1.72485231 | 1.72485231 | **1.72485231** | - | 1.72485231 | 1.72485231 |
| | Mean | 1.72548316 | 1.72548316 | **1.72485807** | - | 1.74321178 | 1.74321178 |
| Hu et al. [17] | Best | - | - | 1.72485084 | - | - | - |
| Coello Coello [18] | Best | 1.74830941 | | | | | |
| | Mean | 1.77197269 | | | | | |
| He et al. [19] | Best | - | - | - | - | 1.724852 | - |
| | Mean | - | - | - | - | 1.749040 | - |
| Foryś et al. [4] | Best | 1.7248 | | | | | |
| de Freitas Vaz et al. [20] | Best | - | - | - | - | 1.814290 | - |

Table 7: Minima found by our optimizer with three constraint-handling methods and three neighbourhood sizes, and by five other authors for the welded beam problem.

### 4.2.3 Tension/compression spring design

For the formulation of this problem, refer to Hu et al. [17] and Coello Coello [18]. The test conditions used in Hu et al. [17] are replicated, and the results are also compared to those of other authors for reference (see Table 8).

### 4.2.4 Himmelblau's nonlinear optimization problem

For the formulation of the problem, refer to Hu et al. [17], Coello Coello [18] and Toscano Pulido et al. [21]. Invalid comparisons can be found in the literature, as the problem formulations differ! Thus, the third terms in the conflict and the constraint functions in Hu et al. [17], Coello Coello [18] and Toscano Pulido et al. [21] are different. It must be noted that not all the formulations of all the problems in all the pa-pers referenced in this work were verified in detail. The settings in Hu et al. [17] were replicated for comparison, and it was assumed that de Freitas Vaz et al. [20] used the same formulation. The results are shown in Table 9 and Table 10.

| TENSION / COMPRESSION SPRING DESIGN | | 1 neighbour | | 2 neighbours | | Global | |
|---|---|---|---|---|---|---|---|
| | | Real | Penalized | Real | Penalized | Real | Penalized |
| Penalization | Best | 0.0126703 | 0.0126703 | 0.0126747 | 0.0126747 | **0.0126652** | **0.0126652** |
| | Mean | 0.0128350 | 0.0128350 | 0.0128103 | 0.0128103 | **0.0127956** | **0.0127956** |
| Preserving feasibility | Best | **0.0126654** | **0.0126654** | 0.0126672 | 0.0126672 | 0.0126700 | 0.0126700 |
| | Mean | **0.0127360** | **0.0127360** | 0.0127507 | 0.0127507 | 0.0128180 | 0.0128180 |
| Bisection | Best | **0.0126652** | **0.0126652** | 0.0126652 | 0.0126652 | 0.0126670 | 0.0126670 |
| | Mean | **0.0126860** | **0.0126860** | 0.0127252 | 0.0127252 | 0.0129549 | 0.0129549 |
| Hu et al. [17] | Best | - | - | **0.0126661** | - | - | - |
| Coello Coello [18] | Best | 0.0127048 | | | | | |
| | Mean | 0.0127692 | | | | | |
| He et al. [19] | Best | - | - | - | - | **0.0126652** | - |
| | Mean | - | - | - | - | 0.0127072 | - |
| de Freitas Vaz et al. [20] | Best | - | - | - | - | 0.0131926 | - |

Table 8: Minima found by our optimizer with three constraint-handling methods and three neighbourhood sizes, and by four other authors for the tension/compression minimum weight string design problem.

| HIMMELBLAU'S PROBLEM | | 1 neighbour | | 2 neighbours | | Global | |
|---|---|---|---|---|---|---|---|
| | | Real | Penalized | Real | Penalized | Real | Penalized |
| Penalization | Best | **-31025.8178** | **-31025.6896** | -31025.8178 | -31025.6896 | -31025.8178 | -31025.6896 |
| | Mean | **-31025.8178** | **-31025.6896** | -31025.8178 | -31025.6896 | -31025.8178 | -31025.6896 |
| Preserving feasibility | Best | -31025.5474 | -31025.5474 | -31025.5614 | -31025.5614 | **-31025.5614** | **-31025.5614** |
| | Mean | -31025.0798 | -31025.0798 | -31025.5548 | -31025.5548 | **-31025.5614** | **-31025.5614** |
| Bisection | Best | **-31025.5614** | **-31025.5614** | -31025.5614 | -31025.5614 | -31025.5614 | -31025.5614 |
| | Mean | **-31025.5614** | **-31025.5614** | -31025.5614 | -31025.5614 | -31022.7249 | -31022.7249 |
| Hu et al. [17] | Best | - | - | **-31025.5614** | - | - | - |
| de Freitas Vaz et al. [20] | Best | - | - | - | - | -31012.1000 | - |

Table 9: Minima found by our optimizer with three constraint-handling methods and three neighbourhood sizes, and by two other authors for the Himmelblau's problem as posed in Hu et al. [17].

| Himmelblau's Problem | $x_1$ | $x_2$ | $x_3$ | $x_4$ | $x_5$ |
|---|---|---|---|---|---|
| Penalization | 77.99997324 | 32.99997741 | 27.07010953 | 45.00001602 | 44.96929014 |
| Preserving feasibility | 78.00000000 | 33.00000000 | 27.07099711 | 45.00000000 | 44.96924255 |
| Bisection | 78.00000000 | 33.00000000 | 27.07099711 | 45.00000000 | 44.96924255 |
| Hu et al. [17] | 78.00000000 | 33.00000000 | 27.07009700 | 45.00000000 | 44.96924255 |
| de Freitas Vaz et al. [20] | 78.0000 | 33.0000 | 27.1106 | 45.0000 | 45.0000 |

Table 10: Coordinates of the minima found by our optimizer and by two other authors for the Himmelblau's problem as posed in Hu et al. [17].

In order to compare our results to those of other authors, the problem is formulated as in Toscano Pulido et al. [21], and the results also compared to those of He et al. [19] and Hu et al. [3] for reference (see Table 11 and Table 12).

| HIMMELBLAU'S PROBLEM | | 1 neighbour | | 2 neighbours | | Global | |
|---|---|---|---|---|---|---|---|
| | | Real | Penalized | Real | Penalized | Real | Penalized |
| Penalization | Best | -30665.953 | -30665.746 | -30665.953 | -30665.746 | -30665.953 | -30665.746 |
| | Mean | -30665.953 | -30665.746 | -30665.953 | -30665.746 | -30665.953 | -30665.746 |
| Preserving feasibility | Best | -30665.538 | -30665.538 | -30665.539 | -30665.539 | -30665.539 | -30665.539 |
| | Mean | -30665.488 | -30665.488 | -30665.539 | -30665.539 | -30665.539 | -30665.539 |
| Bisection | Best | -30665.539 | -30665.539 | -30665.539 | -30665.539 | -30665.539 | -30665.539 |
| | Mean | -30665.539 | -30665.539 | -30665.539 | -30665.539 | -30624.748 | -30624.748 |
| Toscano Pulido et al. [21] | Best | - | - | - | - | -30665.500 | - |
| | Mean | - | - | - | - | -30665.500 | - |
| He et al. [19] | Best | - | - | - | - | -30665.539 | - |
| | Mean | - | - | - | - | -30665.539 | - |
| Hu et al. [3] | Best | | | | | -30665.500 | |
| | Mean | - | - | - | - | -30665.500 | - |

Table 11: Minima found by our optimizer with three constraint-handling methods and three neighbourhood sizes, and by three other authors for the Himmelblau's problem as posed in Toscano Pulido et al. [21].

| Himmelblau's Problem | $x_1$ | $x_2$ | $x_3$ | $x_4$ | $x_5$ |
|---|---|---|---|---|---|
| Penalization | 78.0000 | 33.0000 | 29.9938 | 45.0000 | 36.7766 |
| Preserving feasibility | 78.0000 | 33.0000 | 29.9953 | 45.0000 | 36.7758 |
| Bisection | 78.0000 | 33.0000 | 29.9953 | 45.0000 | 36.7758 |

Table 12: Coordinates of the minima found by our optimizer for the Himmelblau's problem as posed in Toscano Pulido et al. [21].

## 4.3 General comments

The PSO algorithm is a robust, general-purpose optimization method that can cope with very different continuous problems regardless of their features. It can also cope with problems where the variables can take a set of discrete values either by rounding-off the particles trajectories or by the binary version of the algorithm. However, different settings of the coefficients in the particles' velocity update equation considerably affect the dynamics of the swarm. The general-purpose settings used here together with a fully connected neighbourhood topology result in a general-purpose optimizer that performs well on most problems (see Table 1 to Table 12). In some cases, using some local version might conveniently keep diversity for a longer period of time. Our optimizer is flexible, and all the settings can be introduced by the user, including any number of neighbours influencing every particle. As to the constraint-handling techniques, the penalization method results in the best exploration of the search-space because it turns the problem into unconstrained, PSO being an inherently unconstrained optimization method. The drawbacks are the need to tune additional coefficients, and solutions often moderately violating constraints. While self-adaptive penalizations are an active, ongoing line of research, some tolerance is typically acceptable in real-world applications for marginally activated constraints.

## 5 The jetty scheduling problem

Every scheduling problem can be represented by a set of $n$ jobs to be carried out on a set of $m$ machines, where each job consists of a set of operations to be performed on one or more machines. In particular, the job-shop scheduling problem (JSSP) is an NP-hard combinatorial problem consisting of $n$ jobs to be processed through $m$ machines, where each job must be processed on each machine only once, in a particular order, and with no precedence constraints among different job operations. Each machine can handle one job at a time −which cannot be interrupted− and the processing time is fixed and known. The problem consists of finding an optimum schedule, typically associated to the minimum *makespan* (time required to complete all jobs).

If the space of schedules is relatively small, all combinations might be tried and the best one selected (so-called *exhaustive search*). However this becomes infeasible even for small real-world problems, and more intelligent methods that allow cutting off regions of the space of schedules are required (*branch and bound methods*). *Hill-climbing methods* have also been used to handle these problems, where a schedule is randomly generated and successively improved by altering it and keeping the best between the original and the altered schedules. Since these traditional methods get trapped in local optima, modern methods that can escape local optima such as *Tabu Search* and *Simulated Annealing* have been successfully used. More recently, *Genetic Algorithms* (GA) have been reported to find better solutions for scheduling problems. For applications of GAs to scheduling problems, refer to Yamada et al. [22]. In the last few years, the PSO method −originally suitable for continuous problems only− has also been successfully applied to scheduling problems. In most publications, the PSO method undergoes numerous adaptations to be made suitable for such problems (e.g. Xia et al. [23] and Sha et al. [24]). In contrast, the aim here is to keep the method as close to the original version as possible, so as to show its robustness in handling different types of problems with few or no adaptations.

The jetty scheduling problem can be seen as an instance of the general JSSP, which involves the allocation of vessels to the berths at a port along a given period of time so as to minimize the costs incurred through demurrage. The *makespan* minimization might also be desirable as a secondary objective. Our case study consists of a jetty scheduling problem in an oil refinery, where there are numerous constraints that prevent a given vessel to moor up to a given berth (e.g. unfitness of the vessel into the berth due to vessel length or insufficient depth of water; products' unavailability; berth out of order for a period of time; etc.). Thus, the problem is posed as a set of $n$ shipments to be served on a set of $m$ jetties, where a set of berthing operations are to be performed for each shipment served on a given jetty. For the case study considered here, it is assumed that every jetty has the ability to handle all the operations associated to any shipment feasible to be served.

# 6    PSO-based jetty scheduler

Although the objective is to use basically the same optimizer as for the previous benchmark problems, additional code is required to turn the search algorithm into a scheduler. Thus, the jetty scheduling problem is viewed as a search problem on a discrete space, and the PSO equations are transformed in the simplest way possible: the particles' continuous trajectories are rounded off to match the discrete values.

The search-space is limited to its discrete part, and the continuous *Estimated Time of Berthing* (ETB) values are handled deterministically by additional implementations (so far) unrelated to the PSO code. Hence a search-space composed of as many dimensions as shipments are to be scheduled is constructed, where these variables can take any of a set of integer values representing the berths in the port at the refinery. A point in such search-space allocates the shipments that are to be served on each berth, without giving any information about their ETBs, or even about the order in which they are to be served. Therefore, a solution to the problem is given at present by a point in the search-space plus some deterministic rules attached. A future stage in our research consists of considering each berth and its allocated shipments as a travelling salesman problem, so that the order in which they are to be served can be treated as an independent, embedded optimization problem.

Thus, the information regarding shipments and berths is input to the scheduler. This includes fitness information relating vessels and berths; residency time for each vessel on each berth; each shipment's estimated time of arrival (ETA), *laycan window*[3], and demurrage rate. The precedence must be clearly defined at this stage, as the order of the shipments in a berth is deterministically defined. Thus, earlier ETAs award higher precedence to a job. This will be done by an embedded optimization algorithm in a future stage, which would also consider other aspects like the demurrage rate. While the function to be minimized is the total demurrage at present, the *makespan* minimization will be incorporated as a secondary objective in the future.

Our optimizer also allows fixing some vessels to some berths, fixing some vessels' ETBs, and fixing both, while the remaining shipments and berths are scheduled around those constraints. The activation of multi-berth serving is also possible in our optimizer, although this eliminates the chance to embed the optimization algorithm to decide upon the order of the shipments at each berth, since each berth is not independent from the others for multi-berth visits activated. The constraint-handling technique consists of a deterministic repair algorithm, which searches for the closest berth in which the infeasible shipment can be served.

In order to make the comparison fair, a real-world schedule is fed to our PSO-based scheduler, returning a demurrage cost equal to $2.63835 \times 10^9$. Next, the PSO-based scheduler described in the previous section is run using the same coefficients as when optimizing the benchmark problems in previous sections. Since the problem is high-dimensional, a bigger swarm would probably be more convenient. However, the aim is to find a quick solution to this 142-dimensional NP-hard problem. Therefore, the swarm size is kept to 30, a fully connected topology is chosen to favour fast convergence, and the search is performed along 200 time-steps only. The minimum demurrage found equals $1.36016 \times 10^8$, while the average among the demurrages of all 30 particles equals $3.79679 \times 10^8$. Notice that, since a repairing algorithm is used, every particle represents a feasible solution. The search was carried along 8.3 minutes only, using the software *matlab* and a 1.73 GHz Intel Pentium M processor. The evolution of the average and minimum demurrage is shown in Figure 2.

---

[3] It is the time-span in which the vessel has to be served through its completion since its arrival.

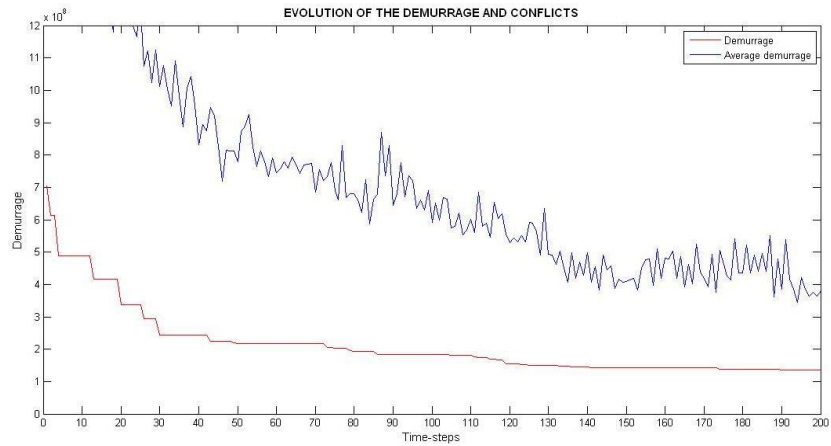

Figure 2: Evolution of the best and average demurrage for a single run of the PSO-based scheduler on a problem with 142 shipments and 10 berths.

# 7 Concluding remarks and future work

The objective to show the robustness of the PSO approach to cope with notably different problems with the same basic algorithm was successfully achieved. In order to strengthen the algorithm's general-purpose characteristics, it is important to use general-purpose settings for the coefficients in the velocity update rule, such as those studied by Innocente [2]. However, not all settings can be kept problem-independent because the conditions of the different problems define different computer power requirements as acceptable. Such settings should keep a balance between exploration and exploitation so as to avoid premature clustering on the one hand, and the lack of fine-tuning or unnecessarily long searches on the other. It appears convenient to use general-purpose settings for the coefficients, and control the exploration/exploitation ratio by setting convenient swarm sizes, maximum number of time-steps permitted, and neighbourhoods' topology. An acceptable number of *fes* always needs to be set in any algorithm, while the topology of the neighbourhood can be set in our optimizer by simply entering the number of neighbours influencing every particle (ring topology). Finally, given an acceptable number of *fes*, the greater the swarm size the more parallel the search −useful in some problems− in detriment of the time-steps assigned for the search to evolve. A swarm of around 30 particles is enough for most problems, but the use of greater swarms should not be plainly disregarded.

Three different constraint-handling methods were implemented: the *penalization*, the *preserving feasibility*, and the *bisection* methods. As much as very good results were obtained, there is room for improvement in each of these techniques. For instance, the self-adaptation of penalization coefficients and the development of techniques to delay the loss of particles' momentum and to generate initial feasible solutions other than brute force are promising lines of research for further improvement.

While binary versions of the method to handle discrete variables comprise promising lines for future research, the algorithm was adapted here to discrete problems by rounding-off the particles' trajectories. Thus, the continuous algorithm was kept virtually unmodified. In turn, the jetty scheduling problem was set as a search problem by generating a discrete space where the variables represent shipments and the discrete values represent the berths where they are to be served. Every solution returned by the PSO algorithm per se consists of a list of shipments to be served on each berth. The order is deterministically decided at present according to the ETAs.

The next steps in our research consist of designing more developed priority rules, incorporating the demurrage rate in addition to the ETAs; testing binary versions of the PSO algorithm; considering other constraint-handling techniques such as the penalization method; including the minimization of the *makespan* as a secondary objective; and considering embedding an independent optimization algorithm on every berth to optimize the order in which the shipments allocated are to be served.

## Acknowledgements

The authors would like to acknowledge the financial and technical support of Maron Systems.